\documentclass[letterpaper]{article}

\usepackage{arxiv}

\usepackage[utf8]{inputenc}
\usepackage[T1]{fontenc}

\usepackage{amsmath,amssymb,amsthm,mathtools,bm}
\usepackage{empheq}
\usepackage{siunitx}
\usepackage{graphicx}
\usepackage{booktabs}
\usepackage{caption}
\usepackage{subcaption}
\usepackage{enumitem}
\usepackage{tikz}
\usetikzlibrary{calc,arrows.meta,angles,quotes,positioning}

\usepackage{url}
\usepackage{microtype}
\usepackage{doi}

\usepackage{algorithm}
\usepackage{algpseudocode}
\usepackage[numbers,sort&compress]{natbib}

\theoremstyle{definition}

\theoremstyle{plain}

\definecolor{teal}{HTML}{0F6E56}
\definecolor{coral}{HTML}{C0451F}
\definecolor{bluec}{HTML}{185FA5}
\definecolor{amberc}{HTML}{BA7517}
\definecolor{purplec}{HTML}{534AB7}
\definecolor{myorange}{HTML}{E8743B}
\definecolor{mygreen}{HTML}{2E9E4F}
\definecolor{mypurple}{HTML}{9B36B8}
\definecolor{mylightblue}{HTML}{9AD0F0}
\definecolor{myblue}{HTML}{1F6FB2}

\title{A Biomimetic Myoelectric Tentacle Prosthesis with Sensorless Object Detection and Vibrotactile Feedback}

\author{
Gabrielle Marion \\
Department of Mechanical Engineering \\
Polytechnique Montréal \\
Montréal, Québec, Canada \\
\texttt{gabrielle.marion@etud.polymtl.ca}
\And
Olivier Lecompte \\
Department of Mechanical Engineering \\
Polytechnique Montréal \\
Montréal, Québec, Canada \\
\texttt{olivier.lecompte@etud.polymtl.ca}
\And
Amandine Gesta \\
Department of Mechanical Engineering \\
Polytechnique Montréal \\
Montréal, Québec, Canada \\
\texttt{amandine.gesta@etud.polymtl.ca}
\And
Parsa Maghsoudloo \\
Department of Mechanical Engineering \\
Polytechnique Montréal \\
Montréal, Québec, Canada \\
\texttt{parsa.maghsoudloo@etud.polymtl.ca}
\And
\href{https://orcid.org/0000-0003-2101-9651}
{\includegraphics[scale=0.06]{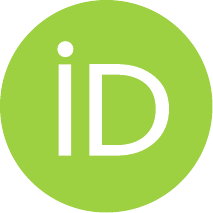}\hspace{1mm}Abolfazl Mohebbi}$^{*}$ \\
Department of Mechanical Engineering \\
Polytechnique Montréal \\
Montréal, Québec, Canada \\
\texttt{abolfazl.mohebbi@polymtl.ca}
}

\renewcommand{\shorttitle}{Biomimetic Myoelectric Tentacle Prosthesis}

\hypersetup{
colorlinks=true,
linkcolor=blue!50!black,
citecolor=blue!50!black,
urlcolor=blue!50!black,
pdftitle={A Biomimetic Myoelectric Tentacle Prosthesis with Sensorless Object Detection and Cumulative Vibrotactile Feedback},
pdfsubject={prosthetics, haptic feedback, myoelectric control, soft robotics},
pdfauthor={Gabrielle Marion, Abolfazl Mohebbi},
pdfkeywords={myoelectric prosthesis, haptic feedback, vibrotactile feedback, sensorless object detection, biomimetic soft robotics},
}

\begin{document}

\maketitle

\renewcommand{\thefootnote}{*}
\footnotetext{Corresponding author. Associate Professor, Director of the Polytechnique Lab for Assistive and Rehabilitation technologies (POLAR), Polytechnique Montréal. Email: \texttt{polar@polymtl.ca}.}
\renewcommand{\thefootnote}{\arabic{footnote}}

\begin{abstract}
This paper presents the design and evaluation of a myoelectric tentacle-shaped prosthesis integrating electromyographic (EMG) control, sensorless object detection, and vibrotactile feedback. The objective was to develop a responsive and intuitive assistive device that adapts to various object shapes while providing sensory feedback to the user. The system relies on EMG signals to control the motion of a flexible, biomimetic structure whose curling geometry follows a logarithmic spiral, enabling it to coil around objects. To ensure stable control, the EMG signal is normalized and filtered, and a threshold-based method identifies user intention. Object contact is detected through a slope-based analysis of motor current, eliminating the need for external sensors, and a haptic feedback strategy based on cumulative vibrotactile stimulation conveys spatial information about the tentacle's configuration. The system was evaluated through quantitative and qualitative tests. The results demonstrate a low response time (\SI{77}{\milli\second} on average), enabling smooth real-time interaction; an object-detection success rate above 90\%, confirming robustness despite EMG variability; and an effective haptic feedback strategy that allowed users to reliably identify the folding zone of the tentacle. The proposed biomimetic design promotes further investigation of expressive artificial limbs by prioritizing expressive functionality over adherence to a predefined, anthropomorphic form factor.
\end{abstract}

\keywords{
myoelectric prosthesis \and
haptic feedback \and
vibrotactile \and
sensorless object detection \and
biomimetic soft robotics
}
\section{Introduction}
\label{sec:intro}
Trauma is the leading cause of upper limb loss, accounting for roughly 80\% of acquired amputations, with most cases occurring in men aged 15 to 45 years~\citep{maduri2019upper}. The event often entails the sudden loss of an active and independent lifestyle. The everyday tasks usually performed effortlessly such as grasping objects, manipulating tools, and performing self-care, suddenly require assistance. This abrupt change frequently contributes to depression, loss of self-esteem, and social isolation, and acceptance of an altered body image is a recognized difficulty during the post-amputation phase~\citep{sansoni2015aesthetic}. Restoring mobility through a prosthetic device is therefore a primary goal, not only to recover independence in daily activities but also to support the self-perception and psychological well-being of amputees.

Depending on user needs, the complexity of the device ranges from a cosmetic prosthesis to a high-technology robotic hand. Three main categories are currently available to patients~\citep{zheng2019priorities, clevelandclinic_prosthetic}: \emph{Passive prosthetic arms} are often preferred when the primary goal is to restore the appearance of the missing limb; while they can be placed in a limited number of configurations, they provide no active degrees of freedom and do not contribute to dexterous tasks. \emph{Body-powered prostheses} offer an alternative for users seeking autonomy. Powered by a pulley system connected to a functional body part, muscle movement is mechanically transmitted to operate terminal devices such as claws or hooks, enabling distal actions such as opening and closing. Although repetitive actions and manual labor can be performed, their functionality remains limited. \emph{Myoelectric prosthetic arms} represent a more advanced solution, operated through electrical signals generated by muscle contraction. By placing electrodes on the skin, these systems detect muscle activity and translate it into movement commands, enabling more intuitive control and improved functionality than traditional prostheses.

Because user needs are numerous and diverse, several factors must be considered in prosthetic design, including dexterity, durability, comfort, and aesthetic appeal. While all contribute to user satisfaction, one of the key elements driving abandonment is the lack of sensory feedback purposely provided by the device. This forces users to rely on visual cues or to exploit the sounds and vibrations emitted by the motors, dedicating their full attention to the grasping task and incurring a significant cognitive load~\citep{papaleo2023integration}. More than 30\% of users stop wearing their prosthesis, citing the lack of haptic feedback as a major contributing factor~\citep{schofield2014applications}. Beyond the functional aspect, haptic feedback facilitates embodiment: rather than feeling like a simple tool, the prosthesis is integrated into the user's body schema, which can improve device acceptance and confidence. To close the control loop, the literature describes several haptic feedback strategies that convey external stimuli to the user through the prosthesis~\citep{schofield2014applications, lecompte2024review}. One is \emph{sensory substitution}, which delivers a feedback signal (e.g., vibration or electrical stimulation) without preserving the original sensory modality of the stimulus detected at the prosthesis. In contrast, when the feedback reproduces the same sensory modality (e.g., pressure felt as pressure), it is referred to as \emph{modality-matched} feedback. Finally, \emph{somatotopic matching} refers to feedback perceived at a body location corresponding anatomically to where the stimulus occurs on the prosthesis.

Restoring the ability to perform activities of daily living (ADLs) is another primary objective of wearing a prosthesis. Defined as the basic tasks necessary for individuals to independently care for themselves~\citep{dollar2014classifying}, ADLs are a critical factor in regaining autonomy and a sense of control over one's mobility. While many studies have investigated user satisfaction with the functional capabilities of prosthetic devices, a frequently reported limitation is the inability of these systems to adapt their grasping configuration to the shape of objects~\citep{cordella2016literature}. As a result, more than 65\% of users report being only moderately satisfied or less with their device. This raises an important question: is the human hand the best tool when it comes to task performance? In this context, the concept of \emph{soft embodiment} in prosthetics offers a promising perspective~\citep{makin2020soft}. Instead of forcing the prosthesis to work exactly like a natural limb, this framework takes advantage of the user's existing motor strategies to control new, sometimes non-anthropomorphic, devices.

Biomimetic approaches are a compelling direction for prosthetic design. By drawing inspiration from mechanisms found in nature, they unlock new strategies to enhance device functionality; soft robotics, for example, takes inspiration from soft-bodied organisms to achieve greater flexibility and adaptability~\citep{varaganti2024recent}. In response to the limited ability of current prostheses to adapt to different grasping configurations, this work presents the design of a myoelectric tentacle-shaped artificial limb whose geometry is inspired by the winding structure of a logarithmic spiral~\citep{wang2025spirobs}. This configuration enables smooth and continuous grasping motions that may improve interaction with objects of varying shapes and sizes. To close the sensory loop between the user and the device, a vibrotactile feedback system is integrated to convey information during the interaction between the prosthesis and the object. The remainder of this paper details the design of the device and its control loop (Section~\ref{sec:methods}), the results of a series of quantitative and qualitative evaluations (Section~\ref{sec:results}), and a discussion of the strengths, limitations, and future directions of the system (Section~\ref{sec:discussion}).

\section{Methods}
\label{sec:methods}

\subsection{Biomimetic Design of the Tentacle}
\label{subsec:design}

The geometry of the prosthesis, designed to interact with objects of varying shapes, inherently requires adaptive grasping strategies. To address this requirement, the prehensile tentacles of octopuses served as the main source of inspiration~\citep{wang2025spirobs}. Their ability to undergo passive deformation illustrates how the characteristics of soft-bodied organisms can be leveraged to develop innovative engineering solutions. While bio-inspiration is a key element of this work, the curling geometry of the tentacle follows the mathematical formulation of a logarithmic spiral.

\begin{figure}[htbp]
  \centering
  \includegraphics[width=0.45\linewidth]{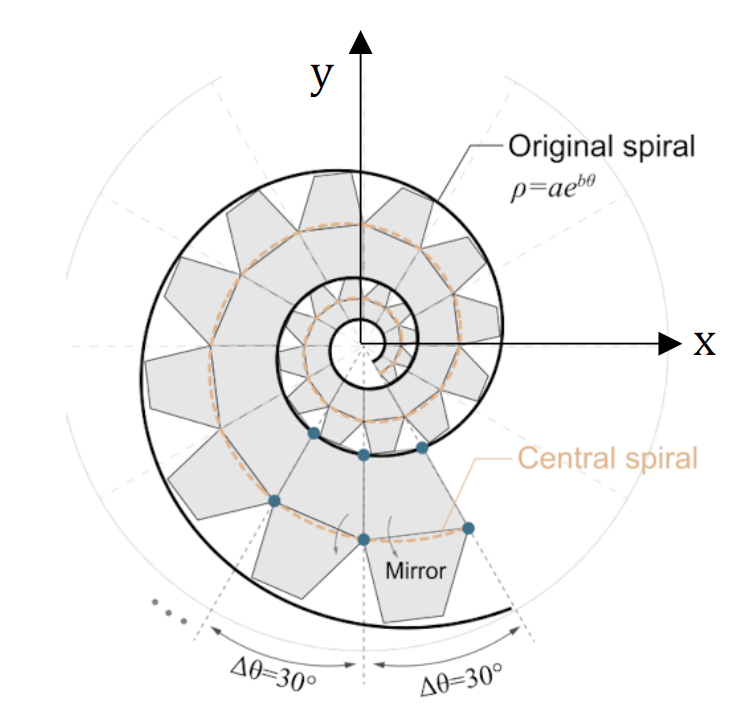}
  \caption{Discretization of the logarithmic spiral into 24 scalable segments. Reproduced from~\citep{wang2025spirobs}.}
  \label{fig:discretization}
\end{figure}

The design is based on a logarithmic spiral discretized into 24 uniformly scaled segments of identical geometry. In polar coordinates, the spiral is defined as
\begin{equation}
  r(\theta) = a \cdot e^{b\theta},
  \label{eq:spiral}
\end{equation}
where $a$ is the initial radius of the spiral, $b$ is the winding rate, $r$ is the radius from the origin, and $\theta$ is the angular position measured from the positive $x$-axis. Figure~\ref{fig:discretization} illustrates how the geometry of the logarithmic spiral is divided into 24 segments.

Once the dimensions of the smallest segment are established, a constant scaling factor is applied to generate each subsequent unit~\citep{wang2025spirobs}. The fixed ratio is defined as
\begin{equation}
  \beta = e^{b\,\Delta\theta},
  \label{eq:ratio}
\end{equation}
where $b$ is the winding rate and $\Delta\theta = \SI{30}{\degree}$ is a fixed discretization step~\citep{wang2025spirobs}. The parameter $b$ is obtained from
\begin{equation}
  \Delta\theta = 2\,\tan^{-1}\!\left(\frac{b\,(e^{2\pi b}-1)}{\sqrt{b^2+1}\,(e^{2\pi b}+1)}\right).
  \label{eq:bparam}
\end{equation}

\begin{figure}[htbp]
    \centering    
    \begin{subfigure}{\linewidth}
        \centering
        \includegraphics[width=0.5\linewidth]{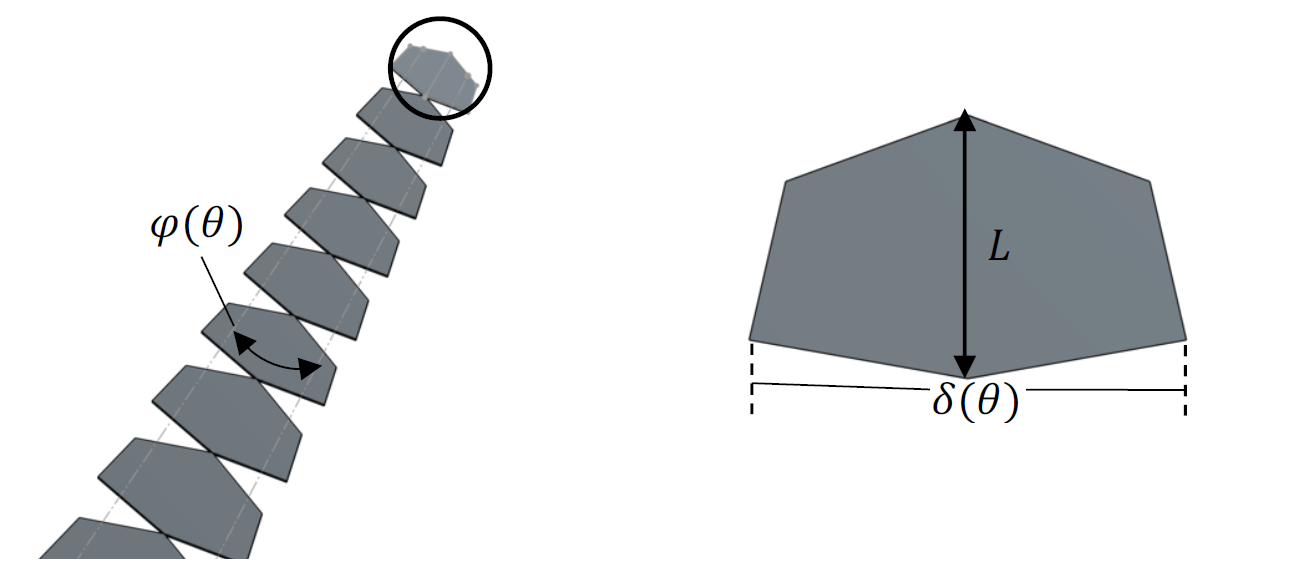} 
        \caption{}
        \label{fig:fig2a}
    \end{subfigure}    
    \vspace{1em} 
    \begin{subfigure}{\linewidth}
        \centering
        \includegraphics[width=0.5\linewidth]{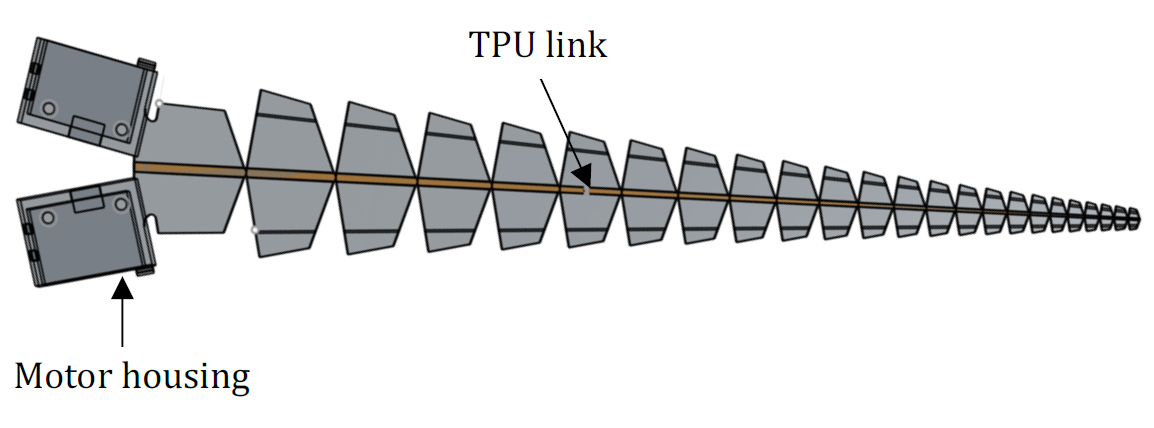} 
        \caption{}
        \label{fig:fig2b}
    \end{subfigure}    
    \caption{(a) Geometric parameters defining the first segment of the tentacle: tapered angle $\varphi(\theta)$, thickness $\delta(\theta)$, and length $L$. (b) CAD model of the \SI{47}{\centi\meter}-long tentacle, showing the central TPU link and the two lateral motor housings.}
    \label{fig:firstsegment-cad}
\end{figure}

The dimensions of the first segment must be defined in order to apply the constant scaling factor of Eq.~\eqref{eq:ratio}. Given a tapered angle $\varphi(\theta) = \SI{10}{\degree}$ chosen for the tentacle, the thickness $\delta(\theta)$ and the length $L$ are the parameters needed to design this first segment (Figure~\ref{fig:firstsegment-cad}a). The thickness is defined as
\begin{equation}
  \delta(\theta) = a\,e^{b(\theta+2\pi)} - a\,e^{b\theta},
  \label{eq:thickness}
\end{equation}
where $b$ is found from Eq.~\eqref{eq:bparam} and the initial radius is fixed to $a = \SI{5}{\milli\meter}$. The length is defined as
\begin{equation}
  L = \frac{a\,\sqrt{b^2+1}}{2b}\,\left(e^{b\,\Delta\theta} - 1\right),
  \label{eq:length}
\end{equation}
with $b$ from Eq.~\eqref{eq:bparam}, $a = \SI{5}{\milli\meter}$, and $\Delta\theta = \SI{30}{\degree}$.

The tapered angle $\varphi(\theta)$ determines the radius of curvature of the tentacle. A smaller tapered angle results in tighter coiling, which limits the prosthesis to grasping small-diameter objects; conversely, higher values produce a more open configuration that limits the ability to generate stable wrapping around objects~\citep{wang2025spirobs}. A value of \SI{10}{\degree} represents a good trade-off, allowing adaptable yet precise grasping.

Based on these parameters, the artificial limb was modeled in an open-access CAD software, to improve accessibility through a low-cost and reproducible fabrication process. The 24 segments were 3D-printed in PLA, selected for its rigidity and cost-effectiveness. At the center of the tentacle is a flexible TPU link that allows the structure to bend smoothly while ensuring that it can return to its original vertical position (Figure~\ref{fig:firstsegment-cad}b). Two housings on either side of the base contain the motors that drive the prosthesis; these motors are actuated by the user's EMG signals, enabling intuitive real-time control.

\subsection{EMG Signal Acquisition and Processing}
\label{subsec:emg}

Because the developed device is a myoelectric prosthesis, the EMG signal serves as the input to the control loop, so accurate acquisition and proper filtering are essential to reliable control. EMG signals are acquired using surface electrodes placed on the biceps muscle, chosen for its accessibility and its ability to generate strong, reliable signals. Following SENIAM recommendations, surface EMG electrodes should be positioned over superficial muscles to ensure signal quality and reproducibility~\citep{hermens2000development}; the biceps is therefore a suitable site for detecting intentional activations in this application. Since the maximum frequency components of the EMG signal are around \SI{500}{\hertz}, the signal is recorded at a sampling rate of \SI{1000}{\hertz}. This is consistent with the Nyquist theorem, which requires the sampling rate to be at least twice the highest frequency present in the signal spectrum, thereby preventing aliasing~\citep{ives2003sampling}. The raw EMG signal is transmitted to an Arduino microcontroller for real-time processing, the purpose of which is to extract a stable control signal from the noisy measurements.

The analog signal read by the Arduino is first centered around zero to facilitate subsequent filtering. A high-pass filter removes the DC offset and attenuates low-frequency motion artifacts, and a low-pass filter reduces high-frequency noise while retaining the relevant frequency content of the EMG signal. A \SI{60}{\hertz} notch filter is then applied to suppress power-line interference. Once filtering is complete, full-wave rectification is performed by taking the absolute value of the filtered signal, converting the bipolar signal into a unipolar one whose amplitude reflects the intensity of muscle activation. Finally, a detection envelope is applied to the rectified signal to produce a more stable output. The processed EMG signal is then transmitted over a serial connection to a computer running a Python-based program.

\subsection{Control Loop}
\label{subsec:control}

The control loop is divided into three modules, each responsible for a function essential to the tentacle's motion: prosthesis actuation via sEMG signals, object detection, and haptic feedback generation. The following subsections describe how these modules operate and interact to provide smooth control of the prosthesis.

\subsubsection{Prosthesis Actuation via sEMG Signals}
\label{subsubsec:actuation}

When using the prosthesis, the first step is to record the user's sEMG activity at rest and then during an effort aiming to achieve maximum voluntary contraction (MVC), a standardization procedure~\citep{fuentesdeltoro2025impact}. The processed EMG signal is normalized to this MVC value, yielding an activation level bounded between 0 and 1. Given the variability of the EMG signal influenced by electrode placement, user physiology, and the ability of each individual to exert force, a robust normalization technique is necessary to obtain comparable performance across users by referencing the signal to a predefined standard. Although several normalization methods exist, MVC was chosen because it has been extensively validated and is considered the gold standard~\citep{fuentesdeltoro2025impact}.

\begin{figure}[htbp]
  \centering
  \includegraphics[width=0.7\linewidth]{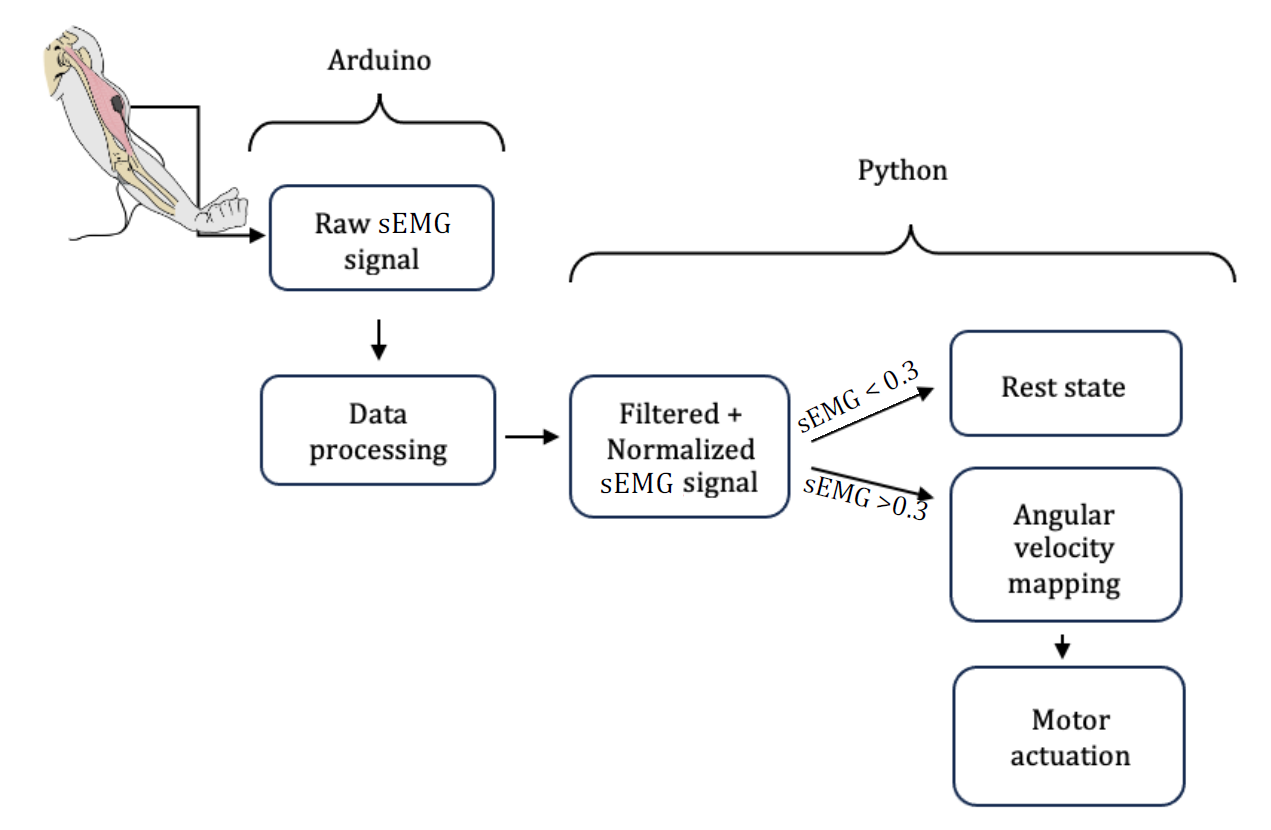}
  \caption{Actuation scheme after MVC normalization. A normalized signal below the \num{0.3} threshold keeps the prosthesis at rest; above it, the activation level is mapped to an angular velocity command sent to the motors.}
  \label{fig:actuation}
\end{figure}

Once calibration is complete, the device is actuated by bilateral servomotors (\emph{Dynamixel XM430-W210-R}) that pull cables routed through either side of the 24 segments to guide the tentacle's motion (Figure~\ref{fig:firstsegment-cad}b). Among the available motor modes (e.g., current-based, position-based), \emph{velocity mode} is the most appropriate for this application: because the normalized EMG signal is used as the control input, this mode readily translates variations in muscle activation into changes in speed, so the motors' rotational speed is controlled according to a proportional relationship. The user can thus adjust the force applied to perform the movement at the desired speed. To prevent unintended motion, the minimum contraction required to generate a rotational command is 30\% of the user's maximum force recorded during MVC normalization. Figure~\ref{fig:actuation} summarizes the actuation scheme after MVC.

\subsubsection{Sensorless Object Detection}
\label{subsubsec:detection}

\begin{figure}[htbp]
    \centering    
    \begin{subfigure}{0.52\textwidth}
        \centering
        \includegraphics[width=\textwidth]{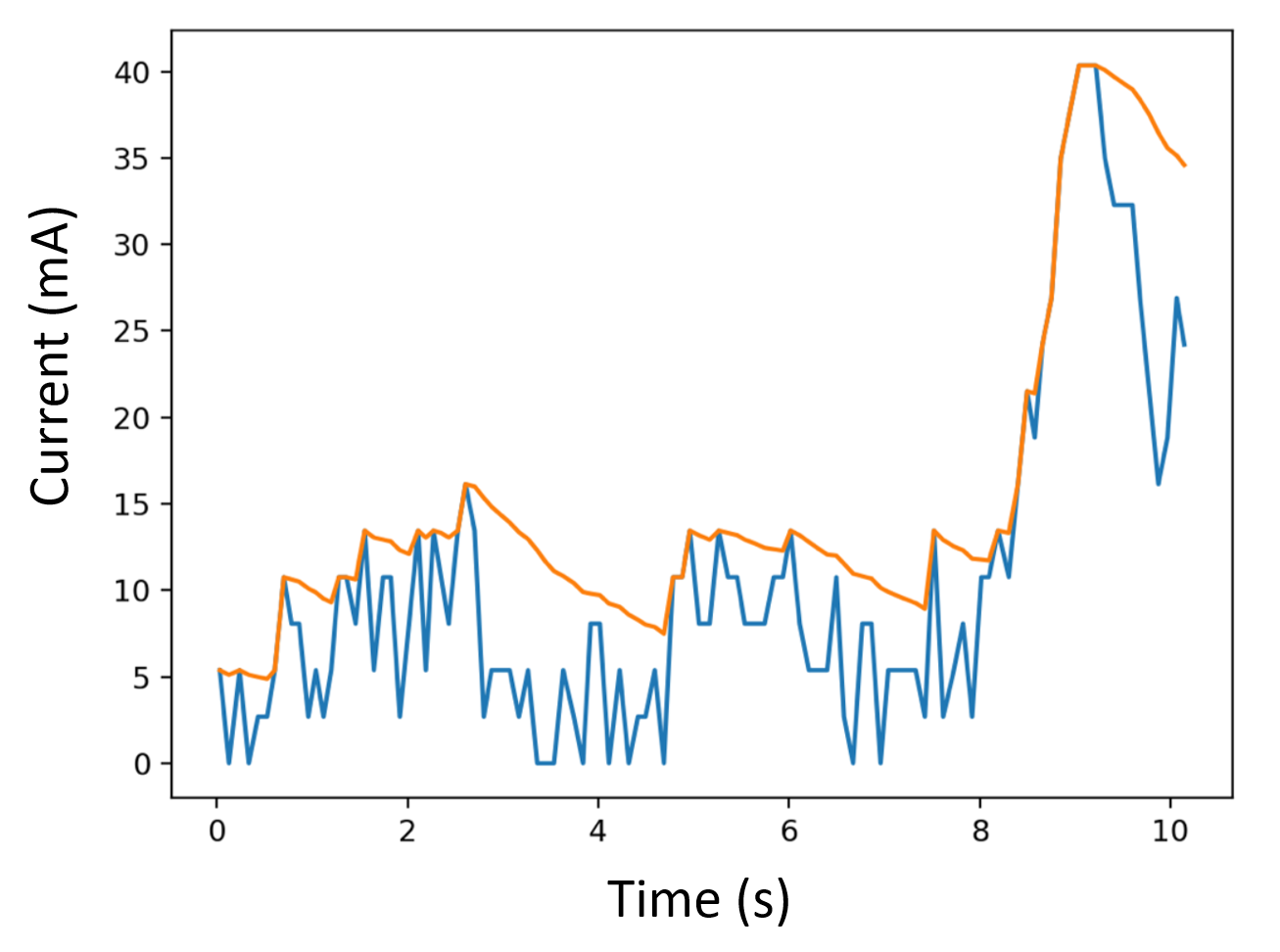} 
        \caption{}
        \label{fig:figcurrent_a}
    \end{subfigure}    
    \begin{subfigure}{0.46\textwidth}
        \centering
        \includegraphics[width=\textwidth]{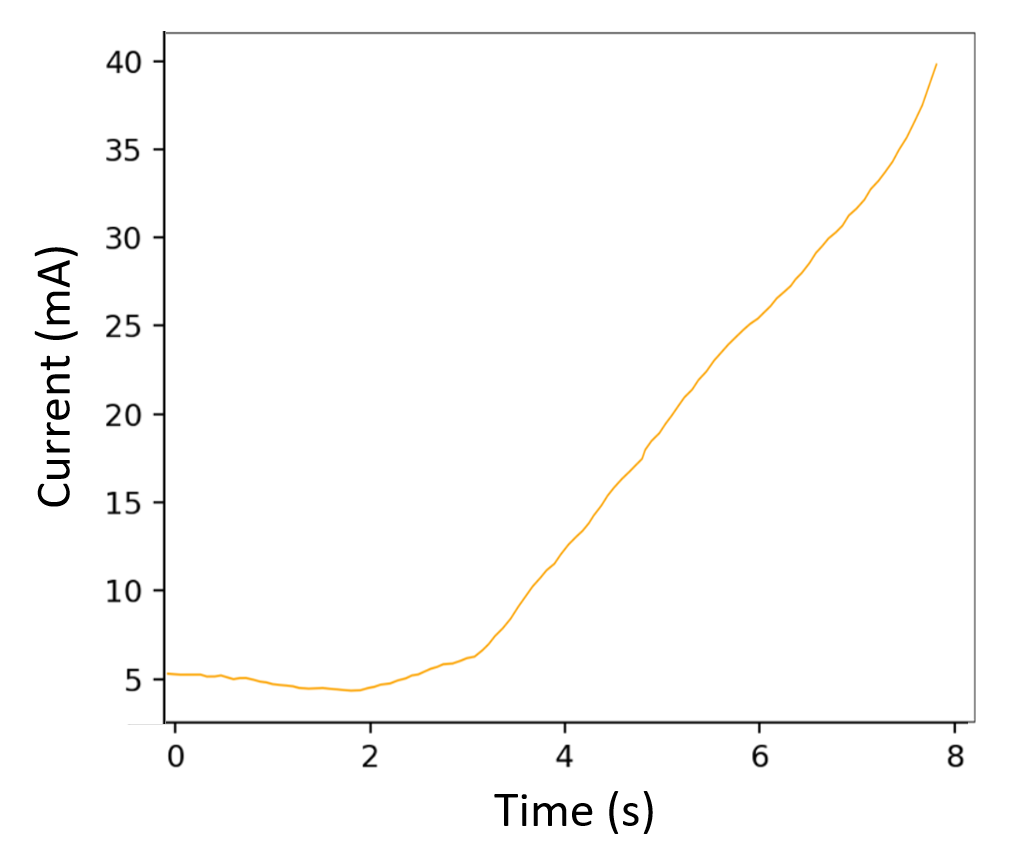} 
        \caption{}
        \label{fig:figcurrent_b}
    \end{subfigure}    
    \caption{(a) Raw motor current values during a grasping trial. (b) Motor current after low-pass filtering. The signal is smoother and exhibits a slope that can be exploited for object detection.}
    \label{fig:current}
\end{figure}

Object detection is the first step in generating haptic feedback, allowing the user to perceive when the tentacle contacts an object. The detection scheme is based on the principle that the motor current increases significantly when the tentacle attempts to wrap around an object: because the motor tries to maintain a constant speed, greater torque is required, which raises the internal current to counteract the encountered resistance. Figure~\ref{fig:current}a shows the raw motor current. At $t \approx \SI{8}{\second}$, an abrupt change in the slope of the current--time curve is visible as the prosthesis grasps an object. The raw signal is noisy, so a low-pass filter is applied to smooth it (Figure~\ref{fig:current}b), yielding a curve whose initial portion exhibits relatively constant current followed by a region of steep increase. This sudden change is exploited to inform the user that the prehension task has been accomplished.

Grasping is detected when the slope of the current curve exceeds an adaptive threshold for more than three consecutive samples. This condition acts as a safety factor, ensuring that detection is not triggered by isolated current spikes but instead reflects a sustained increase corresponding to contact with the object. The adaptive threshold, set empirically after a series of experiments, is given by
\begin{equation}
  \mathrm{Threshold} = \mu + 4\sigma,
  \label{eq:threshold}
\end{equation}
where $\mu$ is the mean current of the baseline and $\sigma$ its standard deviation. An adaptive threshold ensures good reproducibility of performance, as the motor current exhibits significant variability from one experiment to another.

\subsubsection{Haptic Feedback Generation}
\label{subsubsec:haptic}

In the literature, most traditional myoelectric prostheses with a haptic feedback strategy provide information about the grip force exerted on the object~\citep{thomas2023haptic}. The applied force can be measured through load cells or pressure sensors placed at the fingertips, allowing users to modulate their strength to prevent slippage or damage~\citep{abd2022multichannel}.

In this work, because the design is inspired by the flexible and fluid grasping motion of octopus tentacles, its geometry limits the amount of force applied to an object. Unlike conventional grippers or hand-shaped prostheses, the device is not intended to apply strong grasping forces but rather to enable adaptive interaction with the environment. As a result, force-based feedback becomes less relevant, since the prehension task does not typically involve object deformation. Instead, the position of the prosthesis within the $xy$ plane provides meaningful information about task progression, as the object is drawn progressively closer to the user during the movement. For a tentacle-shaped prosthesis, this positional information may be more suitable than grasping force for generating haptic feedback.

\begin{figure}[htbp]
  \centering
  \includegraphics[width=0.5\linewidth]{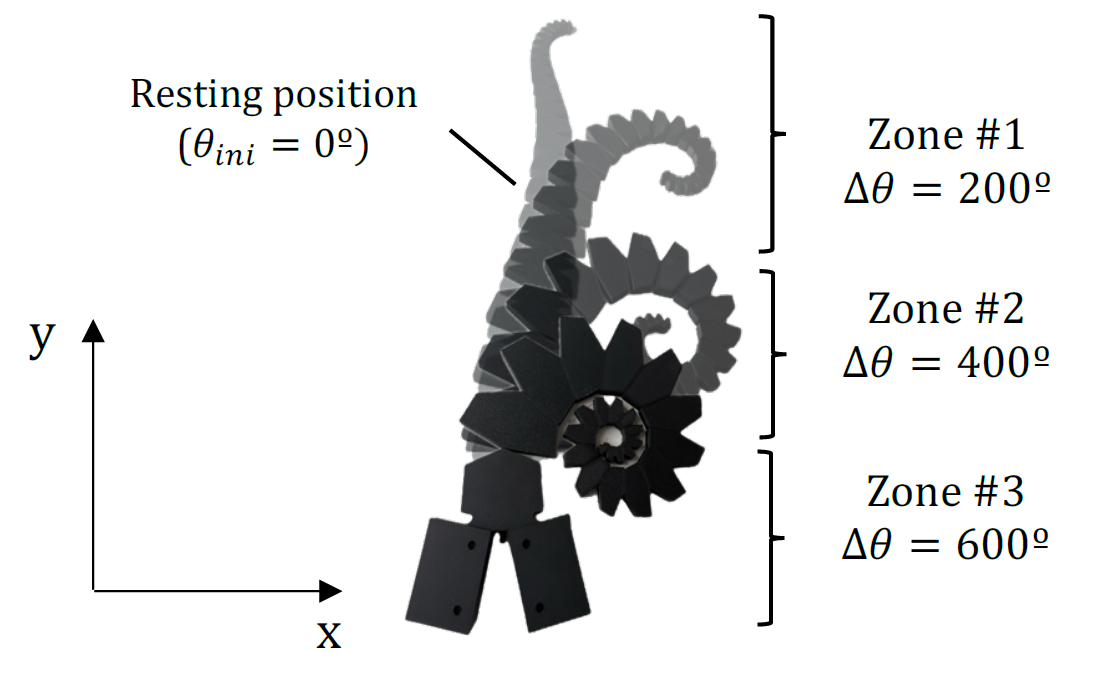}
  \caption{Division of the workspace into three zones from the resting position $\theta_{\mathrm{ini}} = \SI{0}{\degree}$, each spanning an incremental motor rotation of \SI{200}{\degree}.}
  \label{fig:workspace}
\end{figure}

As most frequently reported in the literature, vibrotactile feedback was the chosen modality to convey the position of the prosthesis. To keep the strategy intuitive and easy to learn, the prosthesis workspace in the $xy$ plane was divided into three distinct zones, each corresponding to a motor rotation of $\Delta\theta = \SI{200}{\degree}$ (Figure~\ref{fig:workspace}).

Each zone is associated with a vibrotactile actuator that applies a continuous vibration to the user's forearm. As the tentacle moves and transitions between zones, the corresponding tactor is activated \emph{without} deactivating the one previously engaged. This produces a cumulative increase in vibration, informing the user that the tentacle is coiling and progressing toward its base. Because the motion is controlled by the user's muscle contraction, the task can be ended by stopping at the desired location, accompanied by a distinct vibration sensation. A safety feature stops the motor from rotating once the prosthesis reaches its maximum coiling position, even if the user continues to apply input, preventing damage to the device while limiting the force exerted by the motors.

\subsection{Kinematic Modeling in the XY Plane}
\label{subsec:kinematics}

\begin{figure}[htbp]
    \centering    
    \begin{subfigure}{0.4\textwidth}
        \centering
        \includegraphics[width=\textwidth]{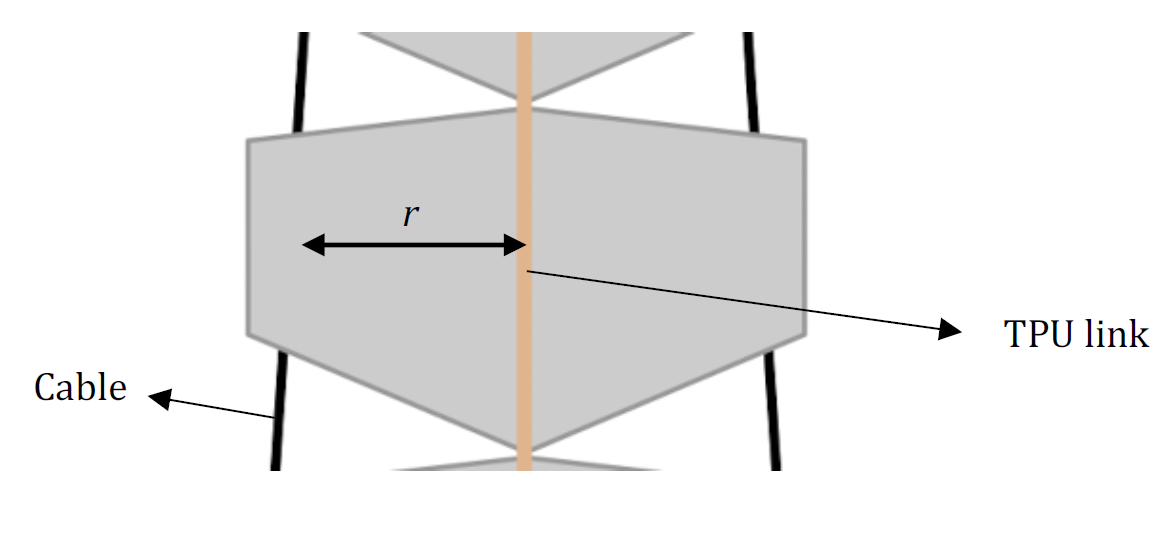} 
        \caption{}
        \label{fig:assumptions_a}
    \end{subfigure}    
    \begin{subfigure}{0.5\textwidth}
        \centering
        \includegraphics[width=\textwidth]{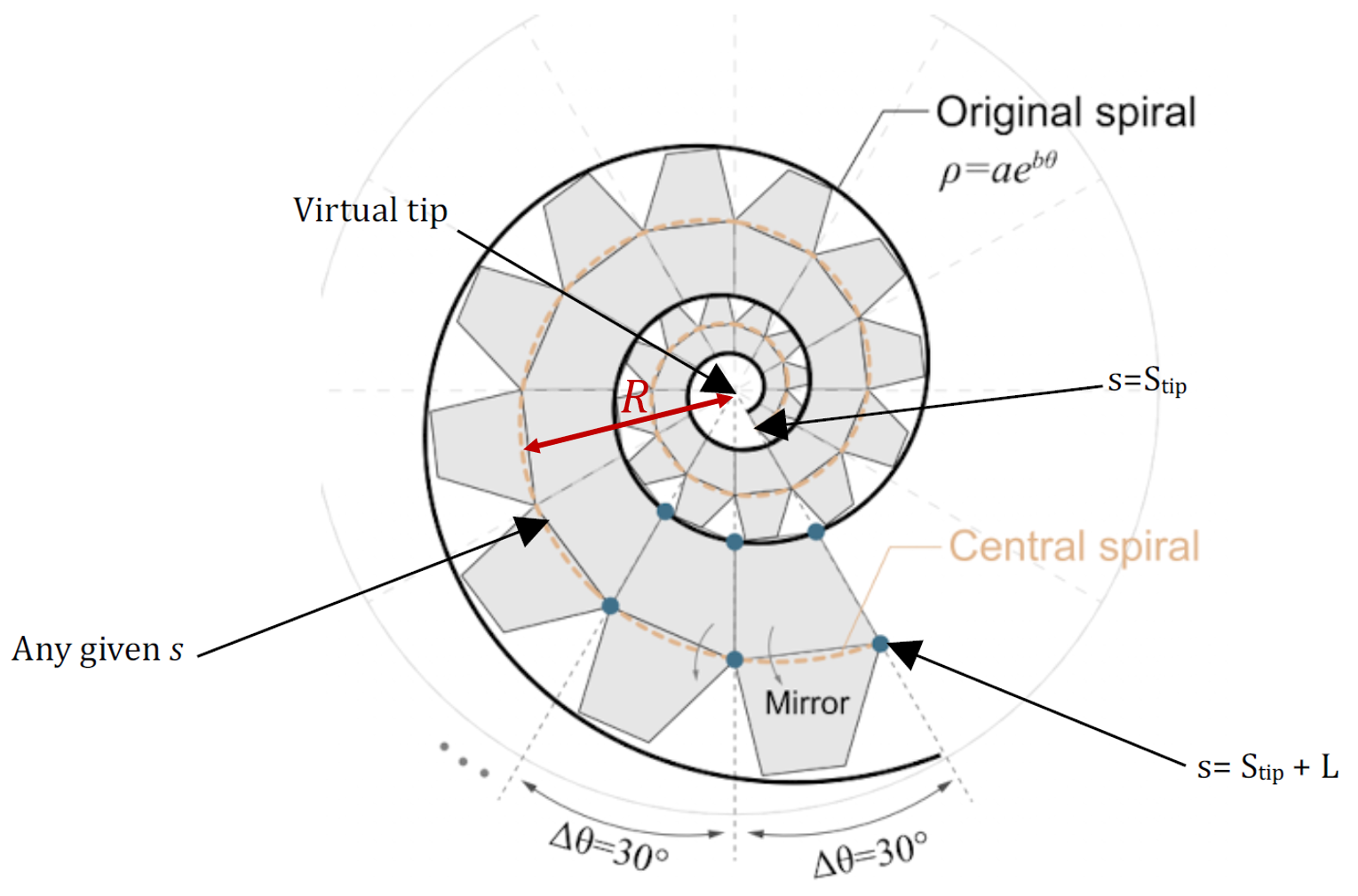} 
        \caption{}
        \label{fig:assumptions_b}
    \end{subfigure}    
    \caption{(a) Modeling assumptions of the tentacle structure and cable configuration, showing the radial cable distance $r$ from the central TPU link. (b) Description of the arc-length parameter $s$ along the central axis of the tentacle. Adapted from~\citep{wang2025spirobs}.}
    \label{fig:assumptions}
\end{figure}

A simplified mathematical model of the tentacle motion in the $xy$ plane was developed to formalize the relationship between the prosthesis configuration and the generated haptic feedback. The model is based on the following assumptions:
\begin{itemize}[nosep]
  \item the prosthesis moves across a flat surface;
  \item the tentacle's central TPU link is a neutral axis whose length remains constant;
  \item the cables run at a radial distance $r$ from the neutral axis;
  \item to simplify the calculations, $r$ is assumed constant along the entire length of the tentacle and is defined as the distance between the TPU link and the cable of the 12\textsuperscript{th} segment (Figure~\ref{fig:assumptions}a).
\end{itemize}

The central equation of the model is
\begin{equation}
  \kappa(s) = \frac{1}{bs},
  \label{eq:curvature}
\end{equation}
where $\kappa$ is the curvature of the tentacle along the central axis, $b$ is the winding rate, and $s$ is the arc length between the virtual tip and any given point on the central axis~\citep{wang2025spirobs}. The virtual tip is a fixed geometric point around which the spiral winds and corresponds to the origin of the system; at the actual tip, $s = S_{\mathrm{tip}}$. Figure~\ref{fig:assumptions}b illustrates the difference between these two parameters. By adjusting the cable length, the arc length between the physical tip and the fixed virtual tip increases or decreases, so $s \in [S_{\mathrm{tip}},\, S_{\mathrm{tip}} + L]$, where $L$ is the total arc length of the tentacle.

\paragraph{Mathematical definition of the curvature $\kappa(s)$.}
The parameterization of the curve, whose starting point corresponds to the virtual tip, is
\begin{equation}
  \vec{r}(s) = \begin{pmatrix} x(s) \\ y(s) \end{pmatrix}.
  \label{eq:param}
\end{equation}
The unit tangent vector, expressed relative to the vertical $y$-axis, is
\begin{equation}
  \vec{t}(s) = \frac{d\vec{r}(s)}{ds} = \begin{pmatrix} \sin\theta(s) \\ \cos\theta(s) \end{pmatrix},
  \label{eq:tangent}
\end{equation}
and the unit normal vector at $\vec{r}(s)$ is
\begin{equation}
  \vec{n}(s) = \begin{pmatrix} \cos\theta(s) \\ -\sin\theta(s) \end{pmatrix}.
  \label{eq:normal}
\end{equation}
Differentiating Eq.~\eqref{eq:tangent} therefore gives
\begin{equation}
  \frac{d\vec{t}}{ds} = \begin{pmatrix} \cos\theta(s)\,\dfrac{d\theta}{ds} \\[6pt] -\sin\theta(s)\,\dfrac{d\theta}{ds} \end{pmatrix}
  = \vec{n}(s)\,\frac{d\theta}{ds}.
  \label{eq:dtds}
\end{equation}
According to the Frenet--Serret formula~\citep{webster2010design},
\begin{equation}
  \vec{t}\,'(s) = \vec{n}(s)\,\kappa(s),
  \label{eq:frenet}
\end{equation}
where $\vec{n}(s)$ is the unit normal vector of Eq.~\eqref{eq:normal} and $\kappa(s)$ is the curvature. Comparing Eqs.~\eqref{eq:dtds} and~\eqref{eq:frenet} yields
\begin{equation}
  \kappa(s) = \frac{d\theta}{ds}.
  \label{eq:kappadthetads}
\end{equation}

\paragraph{Infinitesimal length $dl$ of cable pulled.}
During the motion of the tentacle, cable actuation drives the curling behavior. The cables are pulled differentially; assuming a 1:1 ratio, when one cable is stretched the other shortens. The infinitesimal difference in cable length due to differential actuation is
\begin{equation}
  dl = (R+r)\,d\theta - (R-r)\,d\theta = 2r\,d\theta,
  \label{eq:dl}
\end{equation}
where $R$ is the radius of the central axis from the virtual tip (origin) and $r$ is the distance between the central axis and the cable (Figure~\ref{fig:assumptions}).

Substituting Eq.~\eqref{eq:kappadthetads} into Eq.~\eqref{eq:dl}, the infinitesimal length becomes
\begin{equation}
  dl = 2r\,\kappa(s)\,ds.
  \label{eq:dlkappa}
\end{equation}
Integrating Eq.~\eqref{eq:dlkappa} over the tentacle and using Eq.~\eqref{eq:curvature},
\begin{align}
  \Delta l &= \int_{S_{\mathrm{tip}}}^{S_{\mathrm{tip}}+L} 2r\,\kappa(s)\,ds
           = \int_{S_{\mathrm{tip}}}^{S_{\mathrm{tip}}+L} 2r\cdot\frac{1}{bs}\,ds, \nonumber \\
  \Delta l &= \frac{2r}{b}\,\ln\!\left(\frac{S_{\mathrm{tip}}+L}{S_{\mathrm{tip}}}\right)
           \iff
           S_{\mathrm{tip}}(\Delta l) = \frac{L}{e^{\frac{b\,\Delta l}{2r}} - 1}.
  \label{eq:stip}
\end{align}
For $u \in [0, L]$, the curvature of Eq.~\eqref{eq:curvature} can therefore be expressed as
\begin{equation}
  \kappa(u, \Delta l) = \frac{1}{b\,[\,S_{\mathrm{tip}}(\Delta l) + u\,]},
  \label{eq:kappau}
\end{equation}
where $u$ is any point on the central axis of the tentacle.

The goal is to represent the tentacle in the $xy$ plane. Knowing the angle associated with each configuration allows the coordinates of every point on the curve to be determined. Since $\kappa(u) = d\theta/du \Leftrightarrow d\theta = \kappa(u)\,du$, the angle $\theta(u)$ between the tangent at point $u$ on the central axis and the vertical $y$-axis is
\begin{equation}
  \theta(u) = \int_{0}^{u} \frac{1}{b\,[\,S_{\mathrm{tip}}(\Delta l) + u'\,]}\,du'
            = \frac{1}{b}\,\ln\!\left(\frac{S_{\mathrm{tip}}(\Delta l) + u}{S_{\mathrm{tip}}(\Delta l)}\right).
  \label{eq:theta}
\end{equation}
From the parametric equations, it follows that
\begin{equation}
  \begin{cases}
    \dfrac{dx(u)}{du} = \sin\theta(u) \\[8pt]
    \dfrac{dy(u)}{du} = \cos\theta(u)
  \end{cases}
  \iff
  \begin{cases}
    x(u) = \displaystyle\int_{0}^{u} \sin\theta(u')\,du' \\[8pt]
    y(u) = \displaystyle\int_{0}^{u} \cos\theta(u')\,du'
  \end{cases}
  \label{eq:coords}
\end{equation}
where $x(u)$ and $y(u)$, for $u \in [0, L]$, are the Cartesian coordinates of any given point on the central axis of the tentacle.

\subsection{Experimental Validation}
\label{subsec:validation}

\begin{figure}[htbp]
  \centering
  \includegraphics[width=0.65\linewidth]{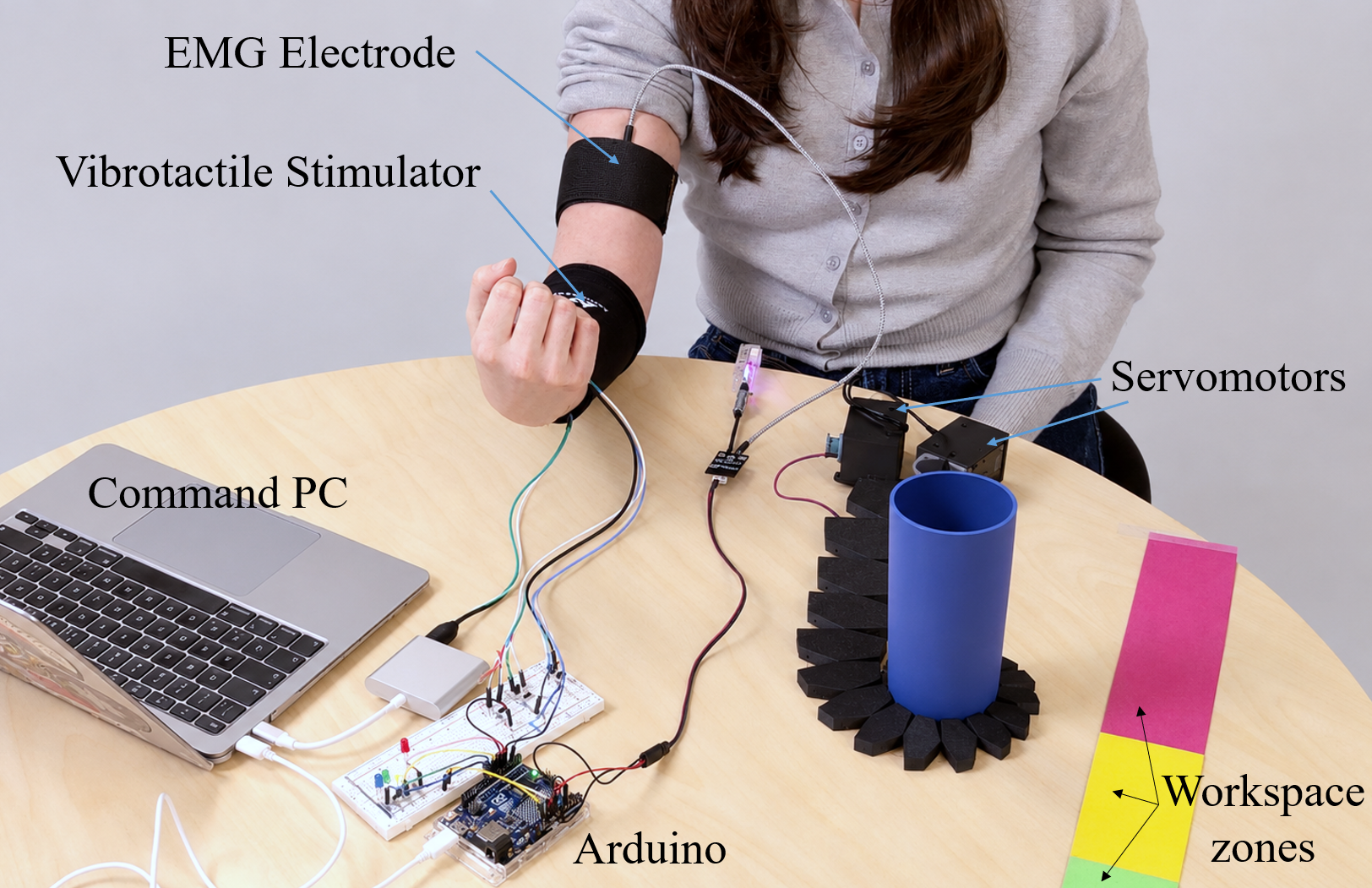}
  \caption{Experimental setup used for validation: surface electrodes on the
  biceps, Arduino acquisition, the tentacle prosthesis and its servomotors,
  the forearm vibrotactile tactors, and the reference LED with the test object.}
  \label{fig:setup}
\end{figure}

The performance of the proposed myoelectric tentacle-shaped prosthesis was evaluated through a series of quantitative and qualitative tests characterizing the responsiveness of the system, the reliability of the object-detection algorithm, and the effectiveness of the haptic feedback strategy. All tests were conducted using the experimental setup shown in Figure~\ref{fig:setup}, comprising the surface EMG electrodes placed on the biceps, the Arduino acquisition stage, the tentacle prosthesis with its bilateral servomotors, the forearm-mounted vibrotactile tactors, and the reference LED together with the test object. 

\paragraph{System response time.}
To evaluate real-time performance, a test measured the response time between the detection of a muscle intention and the activation of the motor. The EMG signal is continuously acquired, then filtered and normalized to reduce noise and obtain a stable signal. A muscle intention is considered detected when the filtered EMG signal exceeds the predefined threshold of 30\% of the MVC value. The response time is measured between the moment of detection and the transmission of the command to the motor and corresponds to the system's overall latency. Several trials were conducted to obtain a representative set of measurements.

\paragraph{Object-detection delay.}
The second test evaluated the delay between the physical contact of the tentacle with an object and the detection of this event by the system. Object detection relies on monitoring the slope of the motor-current signal: when the tentacle encounters resistance, the increased torque produces a measurable rise in current, which is detected when the slope exceeds the adaptive threshold for at least three consecutive samples. The detection delay is the time elapsed between physical contact and the instant at which the detection condition is satisfied.

\paragraph{Haptic feedback effectiveness.}
The third test evaluated the effectiveness of the haptic feedback strategy. The user was asked to identify the folding zone of the tentacle based solely on the vibration perceived on the forearm. Each zone of the tentacle's motion is associated with a vibrotactile actuator, and as the prosthesis coils, the number of active tactors increases, producing a cumulative vibration sensation. To perform the task, the user verbally indicated the zone in which the object was located (zone 1, 2, or 3) based on the perceived feedback. The actual zone was independently tracked by the system and indicated through the activation of an LED whenever the object entered a new zone. This LED served only as a reference for validating the user's response and was not visible to the user, ensuring that identification relied exclusively on haptic perception.

\section{Results}
\label{sec:results}

This section presents the main results concerning the performance of the proposed prosthetic system, including the characteristics of the device, the responsiveness of the control system, the reliability of the object-detection algorithm, and the effectiveness of the haptic feedback strategy.

\subsection{Device Characteristics}
\label{subsec:device}

\begin{figure}[htbp]
  \centering
  \includegraphics[width=0.7\linewidth]{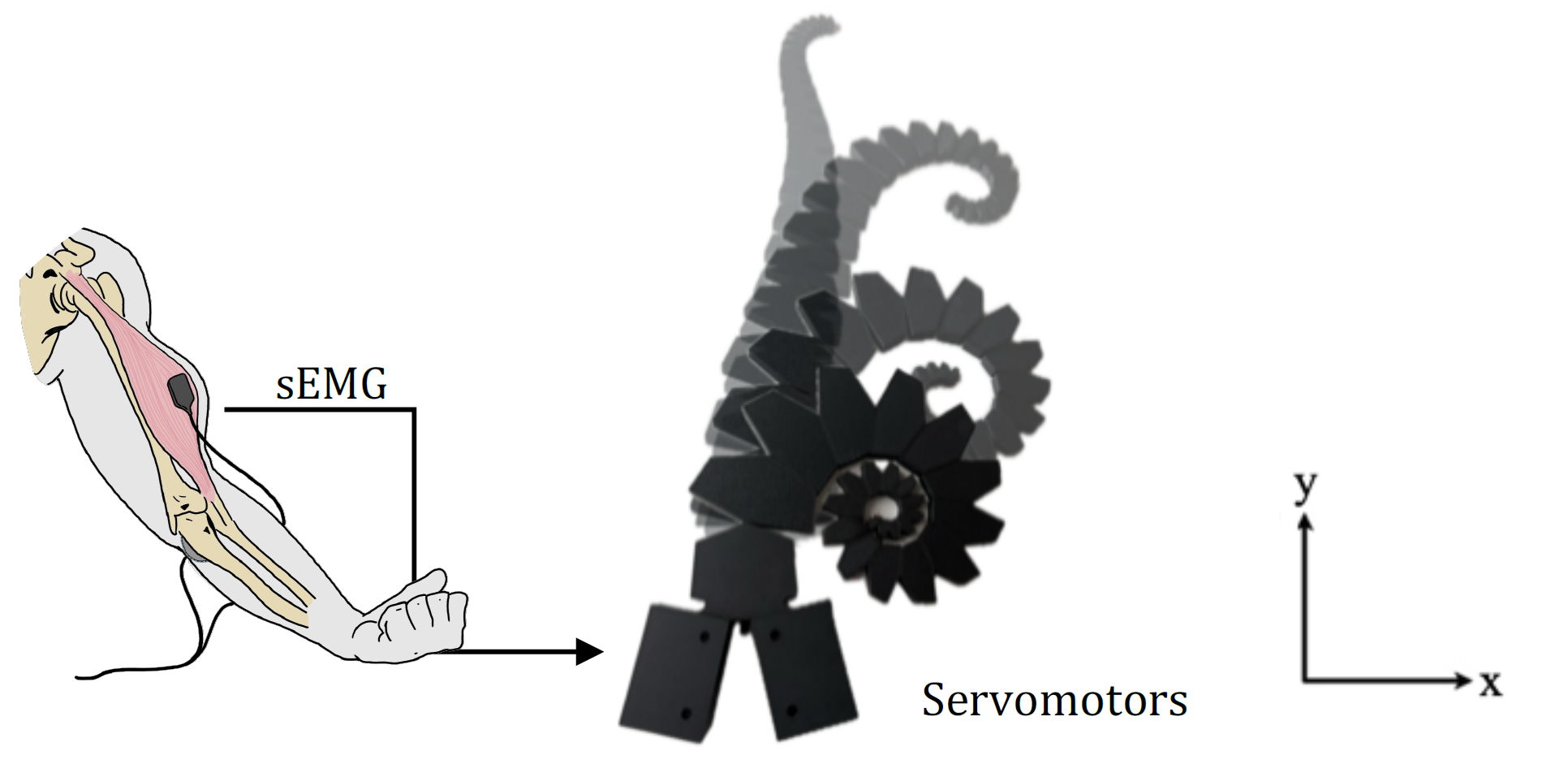}
  \caption{Clockwise movement of the tentacle, from the fully extended position to the maximum winding position, driven by the user's sEMG signal and bilateral servomotors.}
  \label{fig:winding}
\end{figure}

The final device has a total mass of \SI{0.281}{\kilogram}, making it lightweight and suitable for wearable applications. The tentacle consists of 24 segments, for a total length of approximately \SI{47}{\centi\meter}, with a tapered angle fixed at \SI{10}{\degree}. The maximum winding configuration corresponds to approximately two full rotations of the logarithmic spiral. At this position, the prosthesis is capable of grasping objects with a minimum diameter of \SI{1.5}{\centi\meter}, highlighting its ability to interact with small objects. The cable-driven actuation mechanism enables bidirectional motion: pulling one cable releases the other, enabling clockwise and counterclockwise winding of the tentacle, and a return to the fully extended position is possible when the opposite cable is pulled. Figure~\ref{fig:winding} illustrates the clockwise winding of the tentacle as the right cable is pulled to various lengths.

\subsection{System Responsiveness}
\label{subsec:responsiveness}

The response time of the system, defined as the delay between the detection of muscle activation and the initiation of motor movement, was measured to evaluate latency. Over the course of 30 trials, the mean response time was \SI{77}{\milli\second}, with a standard deviation of \SI{19}{\milli\second}. This latency confirms that the overall control pipeline is capable of supporting real-time interaction: the delay between the user's intention and the prosthesis movement was not perceptible during operation. The system responded promptly following muscle activation, contributing to a smooth and intuitive control experience, and this fluidity allowed users to adjust motor speed and decide when to end the movement, giving them a sense of control over their tasks.

\subsection{Object-Detection Reliability}
\label{subsec:detectionresults}

The delay between the physical contact of the tentacle with an object and the detection of this event was measured, with an average detection time of approximately \SI{128}{\milli\second}. Although the low-pass filter introduces a slight temporal shift in the processed signal, this latency is primarily attributable to the requirement that the slope exceed the adaptive threshold for at least three consecutive samples, a condition that keeps the system insensitive to punctual noise spikes.

Despite this delay, the detection algorithm demonstrated a high level of reliability. In all tested scenarios, the system successfully detected object contact in more than 90\% of the trials, indicating that the chosen detection condition represents a suitable trade-off between responsiveness and robustness. Missed detections were mainly attributed to abnormally high slope values during the baseline phase, which resulted in an overestimated threshold and prevented the detection condition from being satisfied upon contact.

The contact-detection tests were performed using a cylindrical object with a diameter of \SI{4.7}{\centi\meter} and a mass of \SI{200}{\gram}. This object was selected because it allowed proper coiling of the tentacle while providing sufficient resistance to generate a measurable increase in motor current. The mass of the object plays a critical role in the detection process: heavier objects induce a greater torque on the motor, producing a more pronounced increase in current that facilitates reliable slope-based detection. Lighter objects, with a mass below \SI{100}{\gram}, could sometimes be detected, but detection was not consistent across trials because the insufficient resistance produced current variations that did not reliably exceed the threshold. To ensure consistent detection, the current implementation is therefore limited to objects with a mass of at least \SI{200}{\gram}.

Because the user controls the tentacle motion during the detection phase, variations in EMG signals can also induce fluctuations in motor current. It was therefore essential for the detection algorithm to remain robust to these variations rather than overly sensitive to changes in signal amplitude. By responding to changes in the current rather than to absolute values, the slope-based mechanism ensures that the system reacts to object contact rather than to current variations caused by user input. The detection success rate indicates that the algorithm can perform reliably under operating conditions involving different levels of user input and signal variability.

\subsection{Haptic Feedback Effectiveness}
\label{subsec:hapticresults}

The effectiveness of the haptic feedback strategy was evaluated through a qualitative test in which the user identified the folding zone of the tentacle based on vibrotactile stimulation. The cumulative activation of vibrotactile actuators produced distinct and progressively stronger vibration patterns, allowing the user to infer the position of the tentacle during the grasping motion. Once the principle of the stimulation had been explained, all participants were able to identify the correct area of the workspace based on the sensation felt in their forearms, indicating that the strategy successfully conveys spatial information in an intuitive manner.

One participant, however, reported difficulty in distinguishing which specific tactor was associated with each zone once activated: the vibration tended to spread across the forearm, making it harder to isolate the pattern associated with each individual zone. This suggests that the user's ability to correctly identify the folding zone was primarily associated with the onset of a new tactor rather than with the steady-state sensation perceived afterward. These results were obtained from more than 30 experiments involving 8 participants.
\section{Discussion}
\label{sec:discussion}

The results presented above indicate that the proposed myoelectric tentacle-shaped prosthesis achieves a high level of performance in terms of responsiveness, reliability, and user interaction. This section discusses the strengths and limitations of the system, as well as directions for future improvement.

\subsection{Responsiveness and Signal Normalization}
\label{subsec:disc_latency}

The measured system latency, with a mean response time of \SI{77}{\milli\second}, confirms that the control pipeline is well suited to real-time applications. This is important in myoelectric control, where delays between user intention and device response degrade usability. From an interaction standpoint, the result falls close to the optimal range reported in the literature for myoelectric prosthesis control, typically between \SI{100}{} and \SI{125}{\milli\second}~\citep{farrell2007optimal}.

The success of prosthesis control also depends on the choice of normalization method, which aims to reduce inter-subject variability and provide a consistent baseline for signal interpretation. While MVC normalization is widely adopted for its simplicity and effectiveness, it presents important limitations~\citep{fuentesdeltoro2025impact}. The approach assumes that users can produce a consistent, maximal contraction, which can be challenging for certain populations: older adults, individuals with neuromuscular disorders, or users experiencing fatigue may not generate reliable MVC values, leading to inaccurate normalization that affects both sensitivity and robustness. An alternative is remote (submaximal) voluntary contraction (RVC), a normalization technique adapted for populations for whom maximal contraction is not feasible~\citep{fuentesdeltoro2025impact}. RVC uses controlled submaximal movements as a reference, such as maintaining a limb against gravity or applying a predefined low force. While it may be better suited to users experiencing muscle fatigue or neuromuscular disorders, it may not capture inter-individual variability in activation levels as effectively as MVC. The choice of normalization therefore depends on the context and the individual using the prosthesis, an important consideration for improving accessibility across a wider range of users.

\subsection{Sensorless Object Detection}
\label{subsec:disc_detection}

Using motor current as a sensing modality for object detection offers a clear advantage in simplicity. By relying on signals already available from the motor, the system avoids additional sensors, reducing hardware complexity and integration effort. The results show that the detection is both robust and reliable, achieving a success rate above 90\%; the slope-based mechanism, combined with the three-consecutive-sample requirement, ensures stable performance.

This approach also introduces limitations. As observed experimentally, detection is highly dependent on the interaction force between the prosthesis and the object. Heavier objects generate sufficient torque to produce a clear current variation, enabling reliable detection, whereas lighter objects do not consistently produce a detectable signal, leading to missed detections. The sensorless approach thus simplifies the system at the cost of reduced sensitivity to low-force interactions. Integrating external sensors, such as pressure or deformation sensors, could provide direct measurements of contact forces and enable detection of lighter objects. However, This can increase complexity through additional hardware, wiring, and signal processing. The choice between sensorless and sensor-based detection must therefore be guided by the intended application and the desired balance between simplicity and performance.

An important strength of the system is its ability to maintain reliable detection despite variations in EMG signals. Because the user directly controls the motor during the detection phase, fluctuations in muscle activation induce variations in motor current. Focusing on changes in the signal rather than its magnitude, the slope-based method is particularly effective at distinguishing user-induced variations from those caused by physical contact. The use of external sensors could also mitigate the effects of EMG variability, making it a promising avenue for improvement.

\subsection{Haptic Feedback}
\label{subsec:disc_haptic}

The haptic feedback test indicates that the proposed vibrotactile strategy is highly effective at conveying spatial information. Participants correctly identified the folding zone of the tentacle, demonstrating that the feedback is both intuitive and reliable, consistent with previous studies showing that vibrotactile feedback can enhance user confidence and performance in grasping tasks~\citep{schofield2014applications}. The use of cumulative activation, in which additional tactors are engaged as the tentacle coils, provides a simple yet effective encoding of spatial information.

Some limitations were also observed. One participant reported difficulty distinguishing between individual tactors once activated, as the vibration tended to spread across the forearm. This suggests that the perceived feedback may depend more on the onset of new stimulation than on steady-state patterns, and it highlights known limitations of vibrotactile feedback, such as poor spatial resolution on the skin and the potential for sensory adaptation, whereby users stop perceiving continuous vibrations over time~\citep{erkat2025improving}. An interesting extension of this work would be to evaluate the temporal perception of haptic feedback; measuring the delay between motor action and perceived vibration could provide insight into how naturally the device is incorporated into the user's perception of their own body.

\subsection{Toward Expressive Prosthetics}
\label{subsec:disc_expressive}

Beyond functional performance, the proposed tentacle-shaped prosthesis aligns with emerging trends in expressive prosthetics. Rather than replicating the appearance of a human limb, expressive prosthetics embrace alternative forms and aesthetics, allowing users to adopt devices that reflect their identity and preferences~\citep{hall2013expressive}. The biomimetic geometry of the prosthesis, combined with its ability to coil around objects, offers functional capabilities that differ from traditional rigid designs, enabling interaction with a wide range of object shapes and opening new possibilities for assistive and supernumerary limb technologies. Studies have shown that most users prioritize functionality, durability, and control over cosmetic appearance~\citep{zheng2019priorities}; several participants even expressed that it would be worthwhile for their prosthesis to stand out, as this could help initiate conversations and reduce the social stigma surrounding amputation. In this context, the proposed design represents not only a technical innovation but also a conceptual shift toward more personalized and expressive assistive devices.

\subsection{Conclusion}
\label{subsec:conclusion}

This work aimed to design and evaluate a myoelectric tentacle-shaped prosthesis capable of providing intuitive control, reliable object interaction, and meaningful haptic feedback. Through a series of quantitative and qualitative validation tests, the system demonstrated strong performance across all key aspects of the control loop.

The measured latency, with a mean response time of approximately \SI{77}{\milli\second}, falls within the optimal range reported in the literature for myoelectric prosthesis control, enabling smooth and natural interaction and allowing users to adjust motor speed and terminate motion with precision. The slope-based detection algorithm, relying on motor-current variations, achieved a success rate exceeding 90\% while eliminating the need for additional sensors, thereby simplifying the system architecture. The haptic feedback strategy, based on cumulative vibrotactile stimulation, enabled participants to reliably identify the folding zone of the tentacle, confirming that the feedback is both intuitive and easy to interpret.

Beyond technical performance, this work highlights the importance of user-centered design in prosthetic development. The tentacle-shaped geometry offers unique functional capabilities, such as adaptation to objects of various shapes, expanding the possibilities of prosthetic interaction beyond traditional rigid designs, while the concept of expressive prosthetics introduces a perspective in which the device is not only functional but also a means of personal expression.

Several limitations remain and provide opportunities for future development. Exploring alternative normalization techniques, such as submaximal reference contractions, could improve accessibility for populations with reduced strength or neuromuscular impairments. Integrating external sensors, such as pressure or deformation sensors, could enhance sensitivity to low-force interactions and enable detection of lighter objects. Finally, measuring the perceived latency of vibrotactile stimulation could clarify how naturally the prosthesis is incorporated into the user's body schema, helping to optimize the timing and design of feedback signals. Overall, the proposed system demonstrates the feasibility of a responsive, reliable, and intuitive myoelectric prosthesis based on a biomimetic tentacle design that prioritizes both performance and user experience by favoring expressive functionality over adherence to a predefined, anthropomorphic form factor.

\bibliographystyle{unsrtnat}
\bibliography{refs}

\end{document}